\documentclass[11pt,a4paper]{article}
\usepackage[hyperref]{emnlp2020}
\usepackage{times}
\usepackage{latexsym}

\usepackage[hang,flushmargin]{footmisc}  % minimize footnote indentation
\usepackage{amsfonts,amsmath,amssymb}
\usepackage[T1]{fontenc}  % fix font related glitches
\usepackage{bold-extra}   % enable \texttt{\textbf{}}
\usepackage{microtype}    % make margins prettier
\usepackage{subfigure}
\usepackage{graphicx}
\usepackage{multirow}
\usepackage{enumitem}

\usepackage[ruled,vlined]{algorithm2e}
\usepackage{url}
\usepackage{bm}           % bold font in math mode

\usepackage{textcomp}
\usepackage{nicefrac}

\aclfinalcopy % Uncomment this line for the final submission
\newcommand{\textsec}[1]{\textsection\ref{#1}}
\newcommand\LN{\linebreak\noindent}  % for micro-management of spacing in line

\newcommand\CRCo{\texttt{CRC1}}
\newcommand\CRCt{\texttt{CRC2}}
\newcommand\CRCr{\texttt{CRC3}}
\newcommand\CRCf{\texttt{CRC4}}
\newcommand\NQ{\texttt{NQ}}
\newcommand\To{\texttt{T1}}
\newcommand\Tt{\texttt{T2}}

\title{Competence-Level Prediction and Resume \& Job Description Matching Using Context-Aware Transformer Models}

\author{Changmao Li$^\spadesuit$, Elaine Fisher$^\diamondsuit$, Rebecca Thomas$^\heartsuit$, \\
  \textbf{Steve Pittard$^\clubsuit$, Vicki Hertzberg$^\diamondsuit$, Jinho D. Choi$^\spadesuit$}\\
  $^\spadesuit$Department of Computer Science, Emory University, Altanta GA, USA \\
  $^\diamondsuit$Nell Hodgson Woodruff School of Nursing, Emory University, Altanta GA, USA \\
  $^\clubsuit$Department of Biostatistics and Bioinformatics, Emory University, Altanta GA, USA \\
  $^\heartsuit$Georgia CTSA Clinical Research Centers, Emory Healthcare, Atlanta GA, USA \\
  {\small \texttt{changmao.li@emory.edu}, \texttt{elaine.fisher@emory.edu}, \texttt{rebecca.s.thomas@emoryhealthcare.org}}\\
  {\small \texttt{wsp@emory.edu}, \texttt{vhertzb@emory.edu}, \texttt{jinho.choi@emory.edu}} }

\date{}

\begin{document}
\maketitle

\begin{abstract}

This paper presents a comprehensive study on resume classification to reduce the time and labor needed to screen an overwhelming number of applications significantly, while improving\LN the selection of suitable candidates. 
A total of 6,492 resumes are extracted from 24,933 job applications for 252 positions designated into four levels of experience for Clinical Research Coordinators (CRC).
Each resume is manually annotated to its most appropriate CRC position by experts through several rounds of triple annotation to establish guidelines. 
As a result, a high Kappa score of 61\% is achieved for inter-annotator agreement.
Given this dataset, novel transformer-based classification models are developed for two tasks: the first task takes a resume and classifies it to a CRC level (\texttt{T1}), and the second task takes both a resume and a job description to apply and predicts if the application is suited to the job (\texttt{T2}).
Our best models using section encoding and multi-head attention decoding give results of 73.3\% to \texttt{T1} and 79.2\% to \texttt{T2}. %, that are very promising.
Our analysis shows that the prediction errors are mostly made among adjacent CRC levels, which are hard for even experts to distinguish, implying the practical value of our models in real HR platforms.

%Moreover, the \texttt{T2} model is able to filter out 27+\% of the applications while retaining 90+\% of good candidates, indicating its practical value in real HR platforms.
\end{abstract}

\section{Introduction}
\label{sec:introduction}

An ongoing challenge for Human Resource (HR) is the process used to screen and match applicants to a target job description with a goal of minimizing recruiting time while maximizing proper matches.
The use of generic job descriptions not clearly stratified by the level of competence or skill sets often leads many candidates to apply every possible job, resulting in misuse of recruiter and applicant's time.
A more challenging aspect is the evaluation of unstructured data such as resumes and CVs, which represents about 80\% of the data processed daily, a task that is typically not an employer's priority given the manual effort involved \cite{darin:19a}.

\noindent The current practice for screening applications involves reviewing individual resumes via traditional approaches, that rely on string/regex matching.
The scope of posted job positions varies by the hiring organization type, job level, focus area, and more.\LN
The latest advent in Natural Language Processing (NLP) enables the large-scale analysis of resumes \cite{deng:18a,myers:19a}. 
NLP models also allow for a comprehensive analyses on resumes and identification of latent concepts that may easily go unnoticed using a general manual process. 
This model's ability to infer core skills and qualifications from resumes can be used to normalize necessary content into standard concepts for matching with stated position requirements \cite{chifu:17a,valdez-almada:18a}. 
However, the task of resume classification has been under-explored due to the lack of resources for individual research labs and the heterogeneous nature of job solicitations.

This paper presents new research that aims to help applicants identify the level of job(s) they are qualified for and to provide recruiters with a rapid way to filter and match the best applicants. 
For this study, resumes submitted to four levels of Clinical Research Coordinator (CRC) positions are used.
To the best of our knowledge, this is the first time that resume classification is explored with levels of competence, not categories.
The contributions of this work are summarized as follows:

\begin{itemize}
\item To create a high-quality dataset that comprises 3,425 resumes annotated with 5 levels of real CRC positions (Section~\ref{sec:dataset}).
\item To present a novel transformer-based classification approach using section encoding and multi-head attention decoding (Section~\ref{sec:approach}).
\item To develop robust NLP models for the tasks of competence-level classification and resume-to-job\_description matching (Section~\ref{sec:experiments}).
\end{itemize}

\begin{table*}[htbp!]
\centering\resizebox{\textwidth}{!}{
\begin{tabular}{c|l}
\bf Type & \multicolumn{1}{|c}{\bf Description} \\
\hline\hline
\multirow{2}{*}{\CRCo}
  & Manage administrative activities associated with the conduct of clinical trials. Maintain data pertaining to research \\
  &  projects, complete source documents/case report forms, and perform data entry. Assist with participant scheduling. \\
\hline
\multirow{3}{*}{\CRCt}
  & Manage research project databases and development study related documents, and complete source documents and \\
  & case report forms. Interface with research participants and study sponsors, determine eligibility, and consent study \\
  & participants according to protocol. \\
\hline
\multirow{4}{*}{\CRCr}
  & Independently manage key aspects of a large clinical trial or all aspects of one or more small trials or research \\ % significant and 
  & projects. Train and provide guidance to less experienced staffs, interface with research participants, and resolve \\
  & issues related to study protocols. Interact with study sponsors, monitor/report SAEs, and resolve study queries. \\
  & Provide leadership in determining, recommending, and implementing improvements to policies and procedures. \\
\hline
\multirow{3}{*}{\CRCf}
  & Function as a team lead to recruit, orient, and supervise research staff. Independently manage the most complex \\
  & research administration activities associated with the conduct of clinical trials. Determine effective strategies for \\
  & promoting/recruiting research participants and retaining participants in long term clinical trials. \\
\end{tabular}}
\caption{Descriptions (and general responsibilities) of the four-levels of CRC positions.}
\label{tab:crc-descriptions}
\end{table*}

\section{Related Work}
\label{sec:related_work}

Limited studies have been conducted on the task of resume classification.
\citet{zaroor_2017} proposed a job-post and resume classification system that integrated knowledge base to match 2K resumes with 10K job posts.
\citet{sayfullina_2017} presented a convolutional neural network (CNN) model to classify 90K job descriptions, 523 resume summaries, and 98 children's dream job descriptions into 27 job categories.
\citet{nasser_2018} hierarchically segmented resumes into sub-domains, especially for technical positions, and developed a CNN model\LN to classify 500 job descriptions and 2K resumes.

Prior studies in this area have focused on classifying resumes or job descriptions into occupational categories (e.g., data scientist, healthcare provider).
However, no work has yet been found to distinguish resumes by levels of competence.
Furthermore, we believe that our work is the first to analyze resumes together with job descriptions to determine whether or not the applicants are suitable for particular jobs, which can significantly reduce the intensive labor performed daily by HR recruiters.

%Transformer-based contextualized embedding approaches such as \texttt{BERT} \cite{devlin_2019}, \texttt{XLNet} \cite{yang_2019a}, \texttt{RoBERTa} \cite{liu_2019}, \texttt{AlBERT} \cite{lan_2019} and \texttt{ELECTRA} \cite{clark_2020} have re-established the state-of-the-art for practically all natural language classification tasks especially the \textsc{GLUE} Dataset \cite{wang_2018}. 
\vspace{-0.2em}
\section{Dataset}
\label{sec:dataset}
\vspace{-0.2em}

%%%%%%%%%%%%%%%%%%%%%%%%%%%%%% Data Collection %%%%%%%%%%%%%%%%%%%%%%%%%%%%%%

\subsection{Data Collection}
\label{ssec:data-collection}

Between April 2018 and May 2019, the department of Human Resources (HR) at Emory University received about 25K applications including resumes in free text for 225 Clinical Research Coordinator (CRC) positions.
A CRC is a clinical research professional whose role is integral to initiating and managing clinical research studies.
There are four levels of CRC positions, \texttt{CRC1-4}, with \CRCf\ having the most expertise.
Table~\ref{tab:crc-descriptions} gives the descriptions about these four CRC levels.
%There are four levels of CRC positions; as the level gets higher, the more professional responsibilities are expected.

\noindent Table~\ref{tab:resume-stats} shows the statistics of the collected applications and the resumes.
Out of the 24,933 applications, 89\% are applied for the entry level positions, \texttt{CRC1-2}, that is expected since \texttt{CRC3-4} positions require more qualifications (\texttt{A}).
At any time, there are various positions posted for the same level from different divisions, cardiology, renal, infectious disease, etc. 
Thus, it is common to see resumes from the same applicant applying to several job postings within the same CRC level.
%Note that each level often makes multiple job postings from different divisions (e.g., Schools of Medicine, Nursing, Public Health) and times so that it is common to see resumes from the same applicant appear in several job postings.

After removing duplicated resumes within the same level, 9,286 resumes remain, discarding 63\% of the original applications (\texttt{B}).
It is common to see the same applicant applying to positions across multiple levels.
After removing duplicated resumes across all levels and retaining only the resumes to the highest level (e.g., if a person applied for both \CRCo\ and \CRCt, retain the resume for only \CRCt), 6,492 resumes are preserved, discarding additional 11\% from the original applications  (\texttt{C}). %, where $45.6/34.7/13.6/6.1\%$ of them apply for \texttt{CRC1$/$2$/$3$/$4}, respectively.

\begin{table}[htbp!]
\vspace{-0.3ex}
\centering\small
\begin{tabular}{l||r|r|r|r||r}
 & \multicolumn{1}{c|}{\textbf{\CRCo}} & \multicolumn{1}{c|}{\textbf{\CRCt}} & \multicolumn{1}{c|}{\textbf{\CRCr}} & \multicolumn{1}{c||}{\textbf{\CRCf}} & \multicolumn{1}{c}{\bf Total} \\ 
\hline \hline
\tt A  & 13,794 & 8,415 & 2,238 & 486 & 24,933 \\ % 11,276 & 6,777 & 1,801 & 241 & 20,095 \\
\tt B  &  4,779 & 3,005 & 1,106 & 396 &  9,286 \\ %  4,271 & 2,560 &   977 & 224 &  8,032 \\
\tt C  &  2,961 & 2,250 &   885 & 396 &  6,492 \\ %  2,733 & 1,947 &   837 & 224 &  5,741 \\
\hline
\tt B$_r$ &  2,730 & 1,702 &   696 & 234 &  5,362 \\
\tt C$_r$ &  1,477 & 1,172 &   542 & 234 &  3,425 \\
\end{tabular}
\caption{The counts of applications (\texttt{A}), unique resumes for each level (\texttt{B}), unique resumes across all levels (\texttt{C}), and resumes from \texttt{B} and \texttt{C} selected for our research while preserving level proportions (\texttt{B}$_r$ and \texttt{C}$_r$).}
\label{tab:resume-stats}
\vspace{-1.2ex}
\end{table}

\noindent For our research, we carefully select 3,425 resumes from \texttt{C} by discarding ones that are not clearly structured (e.g., no section titles) or contain too many characters that cannot be easily converted into text, while keeping similar ratios of the CRC levels (\texttt{C$_r$}).
We also create a set similar to \texttt{B}, say \texttt{B$_r$}, that retains only resumes in \texttt{C$_r$}.
\texttt{C$_r$} and \texttt{B$_r$} are used for our first task (\textsec{ssec:multi-class-model}) and second task (\textsec{ssec:binary-class-model}), respectively.

%These resumes are first preprocessed (Section~\ref{ssec:preprocessing}), annotated (Section~\ref{ssec:annotation}), and then saved in the JSON format so they can be easily read by NLP models.
%43.1/34.2/15.8/6.8

%%%%%%%%%%%%%%%%%%%%%%%%%%%%%% Section Segmentation %%%%%%%%%%%%%%%%%%%%%%%%%%%%%%

\subsection{Preprocessing}
\label{ssec:preprocessing}

The resumes collected by the HR come with several formats (e.g, \texttt{DOC}, \texttt{DOCX}, \texttt{PDF}, \texttt{RTF}).
All resumes are first converted into the unstructured text format, \texttt{TXT}, using publicly available tools.
They are then processed by our custom regular expressions designed to segment different sections in the resumes.
As a results, every resume is segmented into the six sections, \textit{Profile}, \textit{Education}, \textit{Work Experience}, \textit{Activities}, \textit{Skills}, and \textit{Others}.
Table~\ref{tab:section-stats} shows the ratio of resumes in each level including those sections.

\begin{table}[htbp!]
\centering\small
\begin{tabular}{c||c|c|c|c||c}
 & \multicolumn{1}{c|}{\textbf{\CRCo}} & \multicolumn{1}{c|}{\textbf{\CRCt}} & \multicolumn{1}{c|}{\textbf{\CRCr}} & \multicolumn{1}{c||}{\textbf{\CRCf}} & \multicolumn{1}{c}{\bf Total} \\ 
\hline \hline
\tt WoE & 98.0 & 98.3 & 97.2 & 97.4 & 98.0 \\
\tt EDU & 96.0 & 95.6 & 96.3 & 96.6 & 96.0 \\
\tt PRO & 94.4 & 94.3 & 94.1 & 94.0 & 94.3 \\
\tt ACT & 40.4 & 43.4 & 47.4 & 40.2 & 42.5 \\
\tt SKI & 37.7 & 36.4 & 33.6 & 41.5 & 36.9 \\
\tt OTH & 32.2 & 32.8 & 30.8 & 37.2 & 32.5 \\
\end{tabular}
\caption{The existence ratio of each section in the CRC levels. \texttt{WoE}: Work Experience, \texttt{EDU}: Education, \texttt{PRO}: Profile, \texttt{ACT}: Activities, \texttt{SKI}: Skills, \texttt{OTH}: Others.}
\label{tab:section-stats}
\end{table}

\noindent Most resumes consistently include the \textit{Work Experience}, \textit{Education}, and \textit{Profile} sections, whereas the others are often missing.
To ensure the matching quality of our regular expressions, 200 resumes are randomly checked, where 97\% of them are found to have the sections segmented correctly.
Finally, all resumes comprising segmented sections are saved in the \texttt{JSON} format for machine readability.

%%%%%%%%%%%%%%%%%%%%%%%%%%%%%% Annotation %%%%%%%%%%%%%%%%%%%%%%%%%%%%%%

\subsection{Annotation}
\label{ssec:annotation}

2 experts with experience in recruiting applicants for CRC positions of all levels design the annotation guidelines in 5 rounds by labeling each resume with either one of the four CRC levels, \texttt{CRC1-4}, or \textit{Not Qualified} (\NQ), indicating that the applicant is not qualified for any CRC level.
Thus, a total of 5 labels are used for this annotation.
For each round, 50 randomly selected resumes from \texttt{C}$_r$ in Table~\ref{tab:resume-stats}, by keeping similar ratios of the CRC levels as \texttt{C}$_r$, are labeled by those two experts with improvement to subsequent guidelines based on their agreement.

Another batch of 50 resumes are then selected for the next round and annotated based on the revised guidelines.
For batches 2-5, a third person (non-expert) is added and instructed to follow the guidelines developed from prior rounds; thus, annotation is completed by three people for these rounds. 
Table~\ref{tab:ita} shows the Fleiss Kappa scores to estimate the inter-annotator agreement (ITA) for each round with respect to the five competence levels.

\begin{table}[htbp!]
\centering\small
\begin{tabular}{c||r|r|r|r|r}
 & \multicolumn{1}{c|}{\textbf{\texttt{R1}}} & \multicolumn{1}{c|}{\textbf{\texttt{R2}}} & \multicolumn{1}{c|}{\textbf{\texttt{R3}}} & \multicolumn{1}{c|}{\textbf{\texttt{R4}}} & \multicolumn{1}{c}{\textbf{\texttt{R5}}} \\ 
\hline \hline
\NQ     & 13.2 & 53.8 &  38.5 & 52.0 & 66.8 \\ % & 68.2 \\
\CRCo   &  1.3 & 25.0 &  -7.3 & 57.3 & 65.3 \\ % & 55.6 \\
\CRCt   &  9.3 & 39.7 &  41.2 &  5.4 & 33.7 \\ % & 38.9 \\
\CRCr   & 29.1 & 63.5 &  66.7 & 69.8 & 69.6 \\ % & 49.4 \\
\CRCf   & 63.4 & 47.9 & 100.0 &  N/A & -0.7 \\ % & 79.5 \\
\hline
Overall & 16.1 & 45.3 & 40.7 & 55.5 & \bf 60.8 \\ % & 54.7 \\
%\tt A\# & 2 & 3 & 3 & 3 & 3 & 4 \\
\end{tabular}
\caption{Fleiss Kappa scores measured for ITA during the five rounds of guideline development (\texttt{R1-5}). No annotation of \CRCf\ is found in the batch used for \texttt{R4}. The negative kappa scores are achieved for (\CRCo, \texttt{R3}) and (\CRCf, \texttt{R5}) that have too few samples ($\leq 2$).}
\label{tab:ita}
\end{table}

\noindent For \texttt{R1} with no guidelines designed, poor ITA is observed with the kappa score of 16.1\%.
The ITA gradually improves with more rounds, and reaches the kappa score of 60.8\% among 3 annotators, indicating the high quality annotation in our dataset.
The followings give brief summary of the guideline revisions after each round:

\paragraph{Round 1} 
(1) Clarify qualified and not-qualified applicants,
(2) Define transferable skills (e.g, general research experience vs.\ experiences in healthcare),
(3) Define clinical settings, clinical experience, and clinical research experience
(4) Set requirements by levels of academic preparation. % and clinical experience 

\paragraph{Round 2} 
(1) Revise the length of clinical experience based on levels of academic preparation and whether the degree is in a scientific/health related field or non-scientific/non-health related field,
(2) Refine \texttt{CRC2-4} degree requirements, years of clinical research, and clinical experience requirements,
(3) Require clinical research certification for \CRCf.

\paragraph{Round 3} 
(1) Update glossary examples of clinical settings, research experience, and clinical experiences with job titles,
(2) Revise years of experience in clinical roles and research experience. 
(3) Add categorization of foreign trained doctors and bench/laboratory research personnel.

\paragraph{Round 4} 
(1) Remove clinical experience requirements from \texttt{CRC2-4} and require a minimum of 1-year clinical research for those with a scientific vs.\ non-scientific degree,
(2) Revisit laboratory scientist requirements,
(3) Remove academic experience as a research assistant unless it involved over 1000 hours. Rationale: participation by semester is typically data entry or participation in a component of the research but not full engagement in a project. 

\paragraph{Round 5}
Increase the number of years required for a bench/laboratory researcher.

\begin{figure*}[htbp!]
\centering
\includegraphics[scale=0.38]{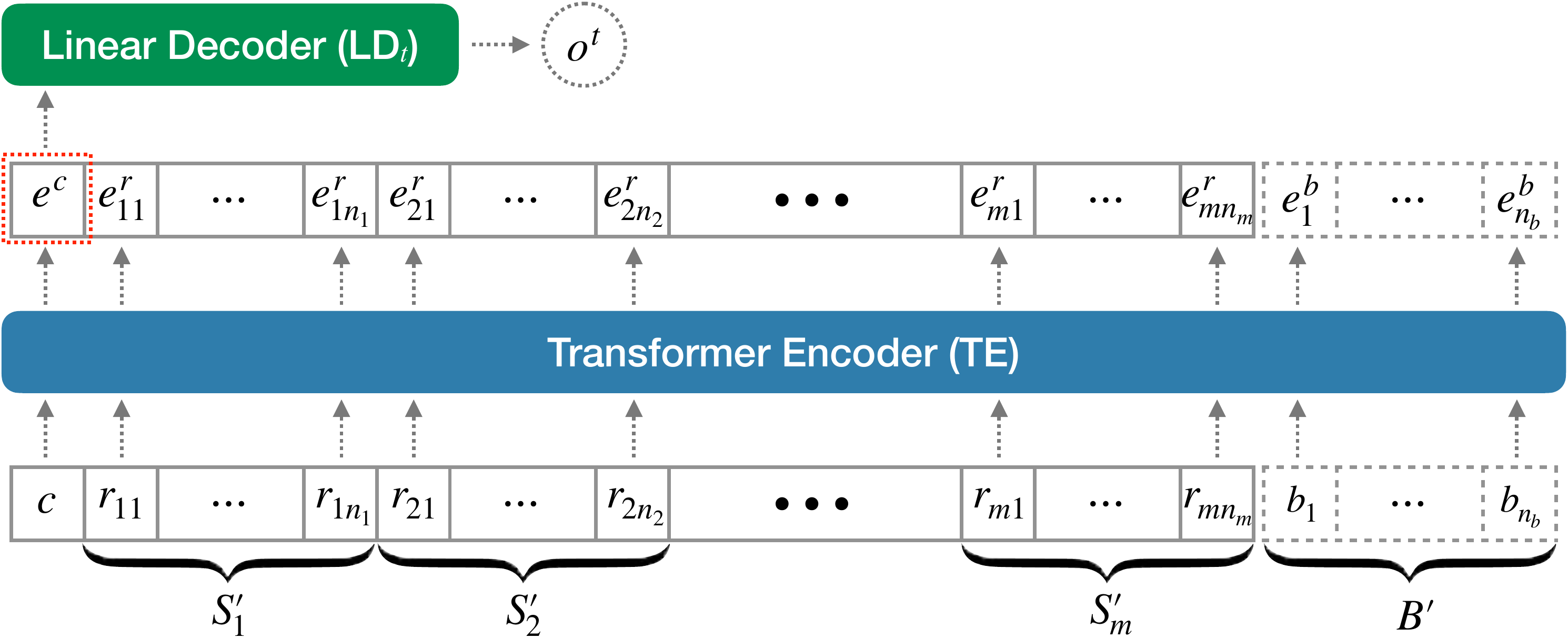}
\caption{The whole context model using section trimming, used as baseline for \To\ (\textsec{sssec:multi-class-section-trimming}) and \Tt\ (\textsec{sssec:binary-class-section-trimming}).}
\label{fig:section-trimming}
\end{figure*}

\noindent During these five rounds, 250 resumes are triple annotated and adjudicated.
Given the established annotation guidelines,\footnote{The annotation guidelines are available at our project page.} additional 3,175 resumes are single annotated and sample-checked.
Thus, a total of 3,425 resumes are annotated for this study.

\section{Approach}
\label{sec:approach}

This section introduces transformer-based neural approaches to address the following two tasks:

\begin{enumerate}
  \item[\To] Given a resume, decide which level of CRC positions that the corresponding applicant is suitable for (Section~\ref{ssec:multi-class-model}).
  \item[\Tt] Given a resume and a CRC job description, decide whether or not the applicant is suitable for that particular job (Section~\ref{ssec:binary-class-model}).
\end{enumerate}

\noindent \To\ is a multiclass classification task where the labels are the five CRC levels including \NQ\ (Table~\ref{tab:ita}).
This task is useful for applicants who may not have clear ideas about what levels they are eligible for, and recruiters who want to match the applicants to the best suitable jobs available to them.

\Tt\ is a binary classification task such that even with the same resume, the label can be either positive (accept) or negative (reject), depending on the job description.
This task is useful for applicants who have good ideas about what CRC levels they fit into but want to determine which particular jobs they should apply to, as well as recruiters who need to quickly screen the applicants for interviews.

%%%%%%%%%%%%%%%%%%%%%%%%%%%%%% Competence-Level Classification %%%%%%%%%%%%%%%%%%%%%%%%%%%%%%

\subsection{Competence-Level Classification}
\label{ssec:multi-class-model}

For the competence-level classification task (\To), a baseline model that treats the whole resume as one document (\textsec{sssec:multi-class-section-trimming}) is compared to context-aware models using section pruning (\textsec{sssec:multi-class-section-pruning}), chunk segmenting (\textsec{sssec:multi-class-chunk-segmenting}), and section encoding (\textsec{sssec:multi-class-section-encoding}).

\subsubsection{Whole-Context: Section Trimming}
\label{sssec:multi-class-section-trimming}

Figure~\ref{fig:section-trimming} shows an overview of the whole context model. 
Let $R=\{S_1,\ldots, S_m\}$ be a resume while $S_i = \{r_{i1}, \ldots, r_{i\ell_i}\}$ is the $i$'th section in $R$ where $r_{ij}$ is the $j$'th token in $S_i$.
Let $N$ be the maximum number of input tokens that a transformer encoder can accept.
Then, $n_i$, the max-number of tokens in $S_i$ allowed to be input, is measured as follows:
\begin{align*}
T   & = \textstyle \sum_{\forall j} |S_j|\\ % = \sum_{\forall j} |\ell_j|\\
n_i & = \min(N, T) \cdot \frac{|S_i|}{T}
\end{align*}
Let $S'_i = \{r_{i1}, \ldots, r_{in_i}\}$ be the trimmed section of $S_i$ by discarding all tokens $r_{ij} \in S_i$ ($n_i < j \leq \ell_i$).
All trimmed sections are appended in order with the special token $c$, representing the entire resume, which creates the input list $I = \{c\} \oplus S'_1 \oplus \cdots \oplus S'_m$.
$I$ is fed into the transformer encoder (\texttt{TE}) that generates the list of embeddings $\{e^c\} \oplus E'_1 \oplus \ldots \oplus E'_m$,\LN where $E'_i = \{e^r_{i1}, \ldots, e^r_{in_i}\}$ is the embedding list of $S'_i$, and $e^c$ is the embeddings of $c$. 
Finally, $e^c$ is fed into the linear decoder (\texttt{LD}$_t$) that generates the output vector $o^t \in \mathbb{R}^d$ to classify $R$ into one of the competence levels (in our case, $d = 5$).

\subsubsection{Context-Aware: Section Pruning}
\label{sssec:multi-class-section-pruning}

\begin{figure*}[htbp!]
\centering
\includegraphics[scale=0.38]{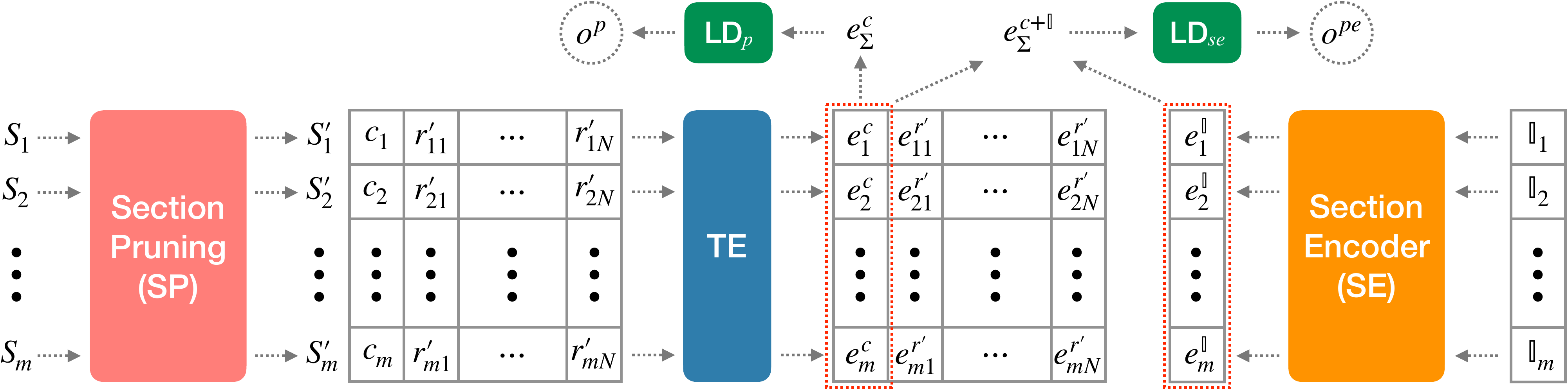}
\caption{The context-aware model using section pruning (\textsec{sssec:multi-class-section-pruning}) and section encoding (\textsec{sssec:multi-class-section-encoding}).}
\label{fig:section-pruning}
\end{figure*}

Section trimming in Section~\ref{sssec:multi-class-section-trimming} allows the whole-context model to take part of every section as input.
However, it is still limited because not all features necessary for the classification are guaranteed to be in the trimmed range.
Moreover, this model makes no distinction between contents from different sections once $S'_{1..m}$ are concatenated.
This section proposes a context-aware model to overcome those two issues by pruning tokens more intelligently and encoding each section separately so that the model learns weights for individual sections to make more informed predictions.
Figure~\ref{fig:section-pruning} shows an overview of the context-aware model using section pruning.

\begin{figure*}[htbp!]
\centering
\includegraphics[scale=0.38]{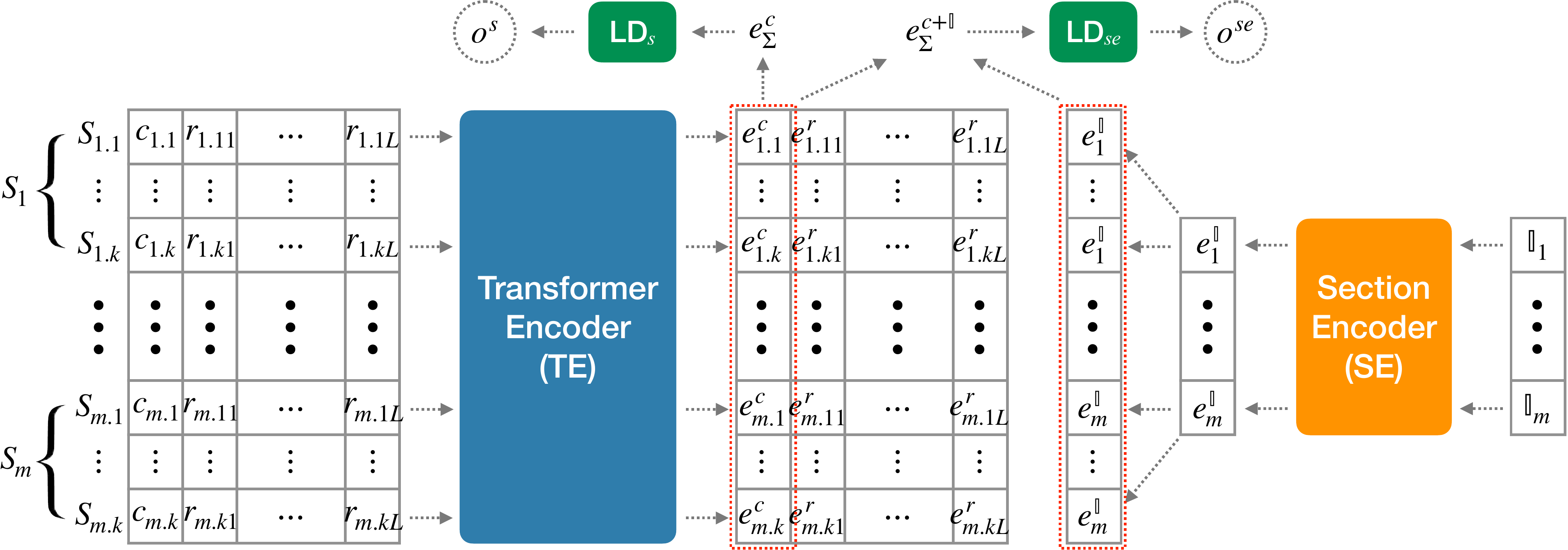}
\caption{The context-aware model using chunk segmenting (\textsec{sssec:multi-class-chunk-segmenting}) and section encoding (\textsec{sssec:multi-class-section-encoding}).}
\label{fig:chunk-segmenting}
\end{figure*}

\noindent Given the maximum number of tokens, $N$, that the transformer encoder (\texttt{TE}) allows, any section $S_i \in R$ that contains more than $N$-number of tokens is pruned by applying the following procedure:

\begin{enumerate}
\item If $|S_i| > N$, remove all stop words in $S_i$.
\item If still $|S_i| > N$, remove all words whose document frequencies are among the top 5\%.
\item If still $|S_i| > N$, remove all words whose document frequencies are among the top 30\%.
\end{enumerate}

\noindent Then, the pruned section $S'_i$ is created for every $S_i$, where $S'_i \subseteq S_i$ and $|S'_i| \leq N$.
Each $S'_i$ is prepended by the special token $c_i$ representing that section and fed into the transformer encoder (\texttt{TE}) that generates the list $\{e_i^c, e^{r'}_{i1}, \ldots, e^{r'}_{iN}\}$, where $e^c$ is the embedding of $c$, called section embedding, and the rest are the embeddings of $S'_i$.
Let $e^{c}_\Sigma = \sum_{i=1}^m e^c_{i}$, that is the sum of all section embeddings representing the whole resume.
Finally, $e^{c}_\Sigma$ is fed into the linear decoder (\texttt{LD}$_p$) that generates the output vector $o^p \in \mathbb{R}^d$ to classify $R$ into a competence level.

\subsubsection{Context-Aware: Chunk Segmenting}
\label{sssec:multi-class-chunk-segmenting}

Section pruning in \textsec{sssec:multi-class-section-pruning} preserves relevant information more than section trimming in \textsec{sssec:multi-class-section-trimming}; however, the model still cannot see the entire resume.
Thus, this section proposes another method that uniformly segments the resume into multiple chunks and encodes each chunk separately.
Figure~\ref{fig:chunk-segmenting} shows the context-aware model using chunk segmenting.
Let $S_i = \{S_{i.1}, \ldots, S_{i.k}\}$ be the $i$'th section in $R$, where $S_{i.j}$ is the $j$'th chunk in $S_i$ and $k = \lceil\nicefrac{|S_{i}|}{L}\rceil$ given the maximum length $L$ of any chunk so that $|S_{i.j}| = L$ for $\forall j < k$ and $|S_{i.k}| \leq L$.\footnote{$S_{i.j} = \{r_{i.j1}, \ldots, r_{i.jL}\}$ and $r_{i.jp}$ is the $p$'th token in $S_{i.j}$, that is $r_{iq} \in S_i$ where $q = L\cdot(j-1) + p$.}
Each chunk $S_{i.j}$ is prepended by the special token $c_{i.j}$ representing that chunk and fed into \texttt{TE} that generates the\LN embedding list $E_{i.j} = \{e_{i.j}^c, e_{i.j1}^r, \ldots, e_{i.jL}^r\}$. % $E_{i.j}$ of $S_{i.j}$.
Let $e^{c}_\Sigma = \sum_{\forall i \forall j} e^c_{i.j}$.
Finally, $e^{c}_\Sigma$ is fed into \texttt{LD}$_s$ that generates the output vector $o^s \in \mathbb{R}^d$ to classify $R$.

\begin{figure*}[htbp!]
\centering
\includegraphics[scale=0.38]{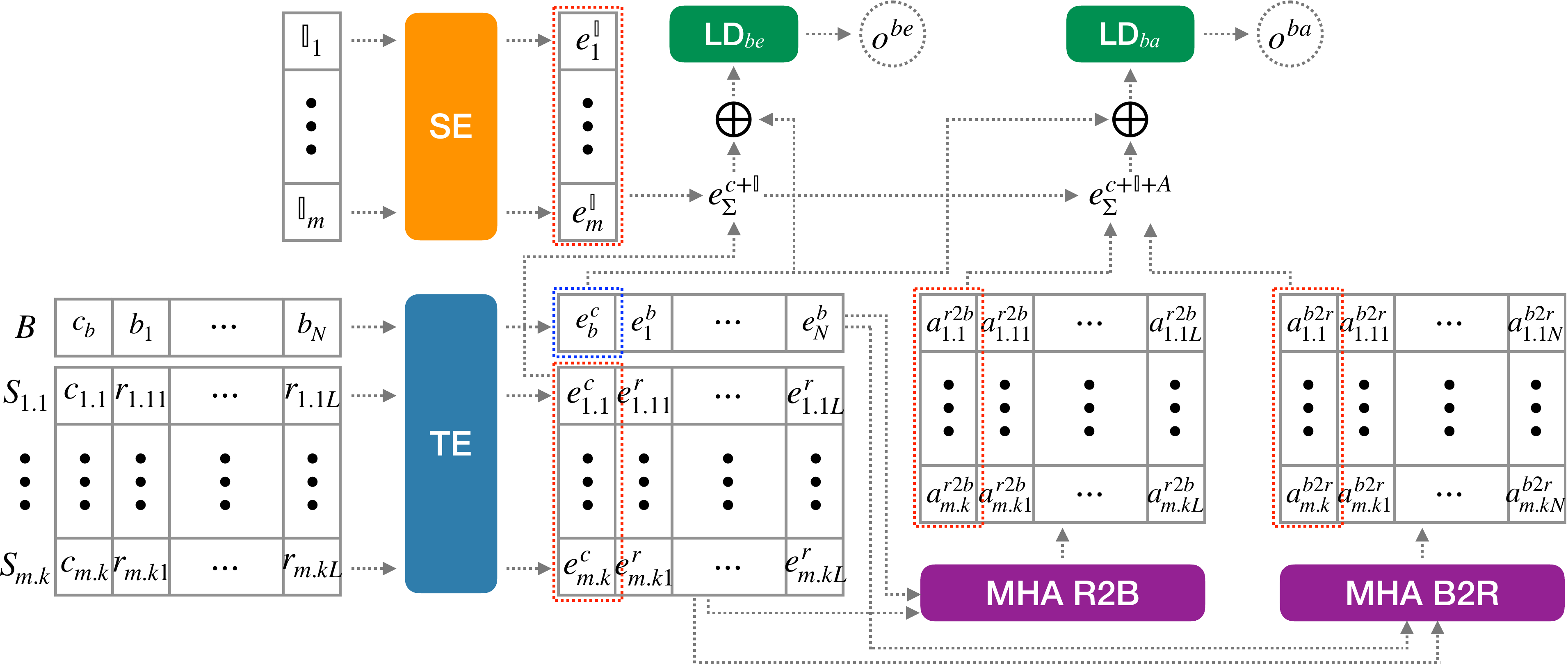}
\caption{The context-aware models using chunk segmenting (\ref{sssec:multi-class-chunk-segmenting}) + section encoding (\textsec{sssec:multi-class-section-encoding}) + job description embedding (\textsec{sssec:binary-class-chunk-encoding}), and multi-head attention between the resume and the job description (\textsec{sssec:binary-class-mha}).}
\label{fig:binary-mha}
\vspace{-0.7ex}
\end{figure*}

\subsubsection{Context-Aware: Section Encoding}
\label{sssec:multi-class-section-encoding}

Chunk segmenting in \textsec{sssec:multi-class-chunk-segmenting} allows the model to see the entire resume; however, it loses information about which sections the chunks belong to.
This section proposes a method to distinctively encode chunks from different sections, that can be applied to both models using section pruning (\textsec{sssec:multi-class-section-pruning}) and chunk segmenting.
Figures \ref{fig:section-pruning} and \ref{fig:chunk-segmenting} describe how section pruning can be applied to those two models.

Let $H = \{\mathbb{I}_1, \ldots, \mathbb{I}_m\}$ be the list of section IDs where $\mathbb{I}_i$ is the ID of the $i$'th section.
$H$ is then fed into the section encoder (\texttt{SE}), an embedding layer that learns the embedding list $E^\mathbb{I} = \{e_1^\mathbb{I}, \ldots, e_m^\mathbb{I}\}$ during training, where $e_i^\mathbb{I}$ is the embedding of $\mathbb{I}_i$.
For the section pruning model in Figure~\ref{fig:section-pruning}, given $E^c = \{e^c_1, \ldots, e^c_m\}$, $F = E^c + E^\mathbb{I} = \{f_1, .., f_m\}$ and $e^{c+\mathbb{I}}_\Sigma = \sum_{i=1}^m f_i$.
For the chunk segmenting model in Figure~\ref{fig:chunk-segmenting}, given $E^c = \{e^c_{1.1}, \ldots, e^c_{m.k}\}$, $F = \{f_{1.1}, \ldots, f_{m.k}\}$ where $f_{i.j} = e^c_{i.j} + e^\mathbb{I}_i$.
Let $e^{c+\mathbb{I}}_\Sigma = \sum_{\forall i\forall j} f_{i.j}$.
Finally, $e^{c+\mathbb{I}}_\Sigma$ is fed into \texttt{LD}$_{se}$ that create $o^{pe} \in \mathbb{R}^d, o^{se} \in \mathbb{R}^d$ for the section pruning and chunk segmenting models, respectively.

%%%%%%%%%%%%%%%%%%%%%%%%%%%%%% Resume-to-Job_Description Matching %%%%%%%%%%%%%%%%%%%%%%%%%%%%%%

\subsection{Resume-to-Job\_Description Matching}
\label{ssec:binary-class-model}

For the resume-to-job\_description matching task (\Tt), the whole-context model is adapted to establish the baseline (\textsec{sssec:binary-class-section-trimming}), and compared to context-aware models using chunk segmenting + section encoding coupled with the job description embedding (\textsec{sssec:binary-class-chunk-encoding}), as well as multi-head attentions between the resume and the job description (\textsec{sssec:binary-class-mha}).

\subsubsection{Whole-Context: Sec./Desc.\ Trimming}
\label{sssec:binary-class-section-trimming}

The whole context model is similar to the one using section trimming in \textsec{sssec:multi-class-section-trimming}
with the additional input from the job description, illustrated as the dotted boxes in Figure~\ref{fig:section-trimming}.
Let $B = \{b_1, \ldots, b_{\ell_b}\}$ be the job description where $b_i$ is the $i$'th token in $B$.
Given the max-number of tokens $N$ that a transformer encoder can accept, the max-numbers of tokens in $S_i$ and $B$, that are $n_i$ and $n_b$ respectively, allowed to be input are measured as followed:
\begin{align*}
T   & = \textstyle \sum_{\forall j} |S_j| + |B| \\
n_i & = \min(N, T) \cdot \nicefrac{|S_i|}{T} \\
n_b & = \min(N, T) \cdot \nicefrac{|B|}{T}
\end{align*}
Let $B' = \{b_1, \ldots, b_{n_b}\}$ be the trimmed job description discarding all tokens $b_j \in B$ ($n_b < j \leq \ell_b$).
Then, the input list $I = \{c\} \oplus S'_1 \oplus\cdots\oplus S'_m \oplus B'$ is created and fed into \texttt{TE} that generates the embedding list $\{e^c\} \oplus E_1 \oplus \ldots \oplus E_m$ $\oplus E_b$, where $E_b$ is the embeddings of $B'$.
Finally, $e^c$ is fed into \texttt{LD} that generates $o^t \in \mathbb{R}^{2}$ to make the binary decision of whether or not $R$ is suitable for $B$.

\subsubsection{Context-Aware: Chunk Segmenting + Section Encoding + Desc. Embedding}
\label{sssec:binary-class-chunk-encoding}

The most advanced competence-level classification model using chunk segmenting (\textsec{sssec:multi-class-chunk-segmenting}) and section encoding (\textsec{sssec:multi-class-section-encoding}) is adapted for the context-aware model with the addition of $B = \{c_b, b_1, \ldots, b_N\}$, which is fed into \texttt{TE} to generate the embedding list $E^b = \{e^c_b, b_1^b, \ldots, b_N^b\}$.
Then, the job description embedding $e^c_b$ is concatenated with the section encoded resume embedding $e^{c+\mathbb{I}}_\Sigma$ (\textsec{sssec:multi-class-section-encoding}) and fed into \texttt{LD}$_{be}$ that generates $o^{be} \in \mathbb{R}^2$.

\subsubsection{Context-Aware: Multi-Head Attention}
\label{sssec:binary-class-mha}

Figure~\ref{fig:binary-mha} depicts an overview of the context-aware model using the techniques in \textsec{sssec:binary-class-chunk-encoding} empowered by multi-head attention \cite{vaswani_2017} 
between\LN the resume $R$ and the job description $B$, which allows the model to learn correlations between individual tokens in $R$ and $B$, $r_*$ and $b_*$, as well as the chunk and job description representations, $c_*$. 

Let $\mathcal{E}^r \in \mathbb{R}^{\gamma \times \lambda}$ be the matrix representing $R$, where $\gamma$ is the total number of chunks across all sections in $R$, $\lambda = L+1$, and $L$ is the max-length of any chunk.
Thus, each row in $\mathcal{E}^r$ is the embedding list $E_{i.j} \in \mathbb{R}^{1 \times \lambda}$ of the corresponding chunk $S_{i,j}$.
Let $\mathcal{E}^b \in \mathbb{R}^{\gamma \times \nu}$ be the matrix representing $B$ where $\nu = N+1$ and $N$ is the max-length of $B$.
Each row in $\mathcal{E}^b$ is a copy of the embedding list $E^b \in \mathbb{R}^{1 \times \nu}$ in \textsec  {sssec:binary-class-chunk-encoding}.
Thus, every row is identical to the other rows in $\mathcal{E}^b$.
These two matrices, $\mathcal{E}^r$ and $\mathcal{E}^b$, are fed\LN into two types of multi-head attention (\texttt{MHA}) layers, one finding correlations from $R$ to $B$ (\texttt{R2B}) and the other from $B$ to $R$ (\texttt{B2R}), which generate two attention matrices, $\mathcal{A}^{r2b} \in \mathbb{R}^{\gamma \times \lambda}$ and $\mathcal{A}^{b2r} \in \mathbb{R}^{\gamma \times \nu}$.

The embeddings of the chunks, $\{e^c_{1.1}, \ldots, e^c_{m.k}\}$, and the section encodings, $\{e_1^\mathbb{I}, \ldots, e_m^\mathbb{I}\}$, as well as\LN the outputs of \texttt{MHA-R2B}, $\{a_{1.1}^{r2b}, \ldots,a_{m.k}^{r2b}\}$, and \texttt{MHA-B2R}, $\{a_{1.1}^{b2r}, \ldots, a_{m.k}^{b2r}\}$, together make $F^a = \{f_{1.1}^a, \ldots, f_{m.k}^a\}$ s.t.\ $f_{i.j} = e^c_{i.j} + e^\mathbb{I}_{i} + a_{i.j}^{r2b} + a_{i.j}^{b2r}$.
Finally, $e^{c+\mathbb{I}+A}_\Sigma = \sum_{\forall i\forall j} f^a_{i.j}$ is fed into \texttt{LD}$_{ba}$ that generates $o^{ba} \in \mathbb{R}^2$ for the binary classification.

\section{Experiments}
\label{sec:experiments}

%%%%%%%%%%%%%%%%%%%%%%%%%%%%%% Data Distributions %%%%%%%%%%%%%%%%%%%%%%%%%%%%%%

\subsection{Data Distributions}
\label{ssec:data-distributions}

Table~\ref{tab:t1-data-split} shows the data split used to develop models for the competence-level classification task (\To).
The annotated data in the row \texttt{C}$_r$ of Table~\ref{tab:resume-stats} are split into the training (\texttt{TRN}), development (\texttt{DEV}) and test (\texttt{TST}) sets with the ratios of 75:10:15 by keeping similar label distributions across all sets. 

\begin{table}[htbp!]
\centering\small
\begin{tabular}{c||r|r|r||r|r}
 & \multicolumn{1}{c|}{\textbf{\texttt{TRN}}} & \multicolumn{1}{c|}{\textbf{\texttt{DEV}}} & \multicolumn{1}{c||}{\textbf{\texttt{TST}}} & \multicolumn{1}{c|}{\textbf{Total}} & \multicolumn{1}{c}{\textbf{Dist.}} \\
\hline \hline
\NQ   &   355 &  48 &  72 &   475 &  13.87\% \\
\CRCo & 1,510 & 202 & 302 & 2,014 &  58.80\% \\
\CRCt &   286 &  38 &  58 &   382 &  11.15\% \\
\CRCr &   392 &  53 &  79 &   524 &  15.30\% \\
\CRCf &    22 &   3 &   5 &    30 &   0.88\% \\
\hline \hline
Total & 2,565 & 344 & 516 & 3,425 & 100.00\%       
\end{tabular}
\caption{Data statistics for the competence-level classification task (\To) in Section~\ref{ssec:multi-class-model}.}
\label{tab:t1-data-split}
\end{table}

\noindent 70\% of the data are annotated with the entry levels, \CRCo\ and \CRCt, that is not surprising since 77.3\% of the applications are submitted for those 2 levels.
The ratio of \CRCf\ is notably lower than the application ratio submitted to that level, 6.8\%, implying that applicants tend to apply to jobs for which they are not qualified. 
13.9\% of the applicants are \NQ; thus, if our model detects even that portion robustly, it can remarkably reduce human labor.

Table~\ref{tab:t2-data-split} shows the data split used for the resume-to-job\_description matching task (\Tt).
The same ratios of 75:10:15 are applied to generate the \texttt{TRN}: \texttt{DEV}:\texttt{TST} sets, respectively.
Note that an applicant can submit resumes to more than one CRC level.
Algorithm~\ref{algortim:splitting} is designed to avoiding any overlapping applicants across datasets while keeping the similar label distributions (Appendix~\ref{ssec:splitting_algortim}).

%ratios of positions applied and annotated results among them to the greatest extent. The training, development and test set ratio is also approximately 0.75:0.1:0.15. 

\begin{table}[htbp!]
\centering\small
\begin{tabular}{c|c||r|r|r||r|r}
\multicolumn{2}{c||}{} & \multicolumn{1}{c|}{\textbf{\texttt{TRN}}} & \multicolumn{1}{c|}{\textbf{\texttt{DEV}}} & \multicolumn{1}{c||}{\textbf{\texttt{TST}}} & \multicolumn{1}{c|}{\textbf{Total}} & \multicolumn{1}{c}{\textbf{Dist.}} \\
\hline\hline
\multirow{2}{*}{\CRCo}
 & \tt Y & 1,279 & 171 & 257 & 1,707 &  31.84\% \\
 & \tt N &   772 & 100 & 151 & 1,023 &  19.08\% \\
\hline
\multirow{2}{*}{\CRCt}
 & \tt Y &   183 &  25 &  38 &   246 &   4.59\% \\
 & \tt N & 1,086 & 148 & 222 & 1,456 &  27.15\% \\
\hline
\multirow{2}{*}{\CRCr}
 & \tt Y &   153 &  21 &  32 &   206 &   3.84\% \\
 & \tt N &   373 &  46 &  71 &   490 &   9.14\% \\
\hline
\multirow{2}{*}{\CRCf}
 & \tt Y &     8 &   0 &   2 &    10 &   0.19\% \\
 & \tt N &   169 &  22 &  33 &   224 &   4.18\% \\ 
\hline\hline
\multicolumn{2}{c||}{\textbf{Total}}  
         & 4,023 & 533 & 806 & 5,362 & 100.00\%
\end{tabular}
\caption{Data statistics for the resume-to-job\_~description matching task (\Tt) in Section~\ref{ssec:binary-class-model}. \texttt{Y}/\texttt{N}: applicants whose applied CRC levels match/do not match our annotated label, respectively.}
\label{tab:t2-data-split}
\vspace{-1ex}
\end{table}

\noindent Out of the 5,362 applications, 40.5\% of them match our annotation of the CRC levels, indicating that less than a half of applications are suitable for the positions they apply.
The number of matches drops significantly for \CRCt; only 14.5\% are found to be suitable according to our labels.
Too few instances are found for \CRCf; only 4.3\% of the applicants applying for this level match our annotation.

%%%%%%%%%%%%%%%%%%%%%%%%%%%%%% Models %%%%%%%%%%%%%%%%%%%%%%%%%%%%%%

\subsection{Models}
\label{ssec:models}

For our experiments, the BERT base model is used as the transformer encoder \cite{devlin_2019} although our approach is not restricted to any particular type of encoder.
The following models are developed for \To\ (Section~\ref{ssec:multi-class-model}):

\begin{itemize} %[leftmargin=*]
\setlength\itemsep{0em}\small
\item \texttt{W}$_r$: Whole context model + section trimming (\textsec{sssec:multi-class-section-trimming})
\item \texttt{P}: Context-aware model + section pruning (\textsec{sssec:multi-class-section-pruning})
\item \texttt{P$\oplus$I}: \texttt{P} + section encoding (\textsec{sssec:multi-class-section-encoding})
\item \texttt{C}: Context-aware model + chunk segmenting (\textsec{sssec:multi-class-chunk-segmenting})
\item \texttt{C$\oplus$I}: \texttt{C} + section encoding (\textsec{sssec:multi-class-section-encoding})
\end{itemize}

\noindent The followings are developed for \Tt\ (Section~\ref{ssec:binary-class-model}):

\begin{itemize} %[leftmargin=*]
\setlength\itemsep{0em}\small
\item \texttt{W$_{r+b}$}: Whole context + sec./job\_desc.\ trimming (\textsec{sssec:binary-class-section-trimming})

\item \texttt{P$\oplus$I$\oplus$J}: \texttt{P$\oplus$I} + job\_desc.\ embedding ($\approx$\textsec{sssec:binary-class-chunk-encoding})
\item \texttt{P$\oplus$I$\oplus$J$\oplus$A}: \texttt{P$\oplus$I$\oplus$J} + multi-head attention ($\approx$\textsec{sssec:binary-class-mha})
\item \texttt{P$\oplus$I$\oplus$J$\oplus$A$\ominus$E}: \texttt{P$\oplus$I$\oplus$J} - $E^c$  (\textsec{sssec:multi-class-section-encoding})

\item \texttt{C$\oplus$I$\oplus$J}: \texttt{C$\oplus$I} + job\_desc.\ embedding (\textsec{sssec:binary-class-chunk-encoding})
\item \texttt{C$\oplus$I$\oplus$J$\oplus$A}: \texttt{C$\oplus$I$\oplus$J} + multi-head attention (\textsec{sssec:binary-class-mha})
\item \texttt{C$\oplus$I$\oplus$J$\oplus$A$\ominus$E}: \texttt{C$\oplus$I$\oplus$J} - $E^c$  (\textsec{sssec:multi-class-section-encoding})
\end{itemize}

\noindent The \texttt{P$\oplus$I$\oplus$J} model adapts section pruning to generate $e^{c+\mathbb{I}}_\Sigma$ instead of chunk segmenting in \textsec{sssec:binary-class-chunk-encoding}.
For the \texttt{P$\oplus$I$\oplus$J$\oplus$A} model, the attention matrices in \textsec{sssec:binary-class-mha} are reconfigured as $\mathcal{A}^{r2b}, \mathcal{A}^{b2r} \in \mathbb{R}^{m \times \nu}$ ($m$: the number of sections in $R$).
These models are developed to make comparisons between those two approaches for \Tt.
Also, the \texttt{*$\ominus$E} models exclude the embedding list $E^c$ such that $f_{i.j}$ is redefined as $f_{i.j} = e^\mathbb{I}_{i} + a_{i.j}^{r2b} + a_{i.j}^{b2r}$ in \textsec{sssec:binary-class-mha} to estimate the pure impact of multi-head attention.

%%%%%%%%%%%%%%%%%%%%%%%%%%%%%% Results %%%%%%%%%%%%%%%%%%%%%%%%%%%%%%

\subsection{Results}
\label{ssec:results}

Labeling accuracy is used as the evaluation metric for all our experiments.
Each model is developed three times and their average score as well as the standard deviation are reported.\footnote{Appdendix~\ref{ssec:experimental-settings} provides details of our experimental settings for the replicability of this work.} 
Table~\ref{table:multiclass_task_results} shows the results for \To\ achieved by the models in Sec.~\ref{ssec:models}. 
All context-aware models without section encoding perform significantly better, 1.5\% with section pruning (\texttt{P}) and 3.3\% with chunk segmenting (\texttt{C}), than the baseline model (\texttt{W}$_r$).
\texttt{C} shows a greater improvement of 1.8\% than \texttt{P}, implying that the additional context used in \texttt{C} is essential for this task.\LN
Section encoding (\texttt{I}) helps both \texttt{P} and \texttt{C}.
As the result, \texttt{C$\oplus$I} shows 4.2\% improvement over \texttt{W$_r$} and also gives the least variance of 0.16.

\begin{table}[htbp!]
\centering\small
\begin{tabular}{l||c|c|c}
 & \multicolumn{1}{c|}{\textbf{\texttt{DEV}}} & \multicolumn{1}{c|}{\textbf{\texttt{TST}}} & \bm{$\delta$} \\ 
\hline \hline
\texttt{W$_{r}$}    & 69.38 ($\pm$0.14) & 69.06 ($\pm$1.56) & - \\ 
\hline
\texttt{P}          & 68.99 ($\pm$0.49) & 70.58 ($\pm$0.38) & 1.52 \\
\texttt{P$\oplus$I} & 69.19 ($\pm$0.63) & 70.87 ($\pm$0.40) & 1.81  \\ 
\hline
\texttt{C}          & 70.36 ($\pm$0.34) & 72.35 ($\pm$0.24) & 3.29 \\
\texttt{C$\oplus$I} & \textbf{70.64} ($\pm$0.41) & \textbf{73.26} ($\pm$0.16) & \textbf{4.20} \\
\end{tabular}
\caption{Accuracy ($\pm$ standard deviation) on the development (\texttt{DEV}) and test (\texttt{TST}) sets for \To, achieved by the models in Section~\ref{ssec:models}. $\delta$: delta over \texttt{W$_r$} on \texttt{TST}.}
\label{table:multiclass_task_results}
\end{table}

\noindent Table~\ref{table:binary_class_task_results} shows the results for \Tt\ achieved by the models in Section~\ref{ssec:models}.
Neither the context-aware model using section pruning (\texttt{P}) or chunk segmenting  (\texttt{C}) with section encoding (\texttt{$\oplus$I}) performs better than the baseline model (\texttt{W}$_{r+b}$) by simply concatenating the job description embedding (\texttt{$\oplus$J}).
Indeed,\LN none of the \texttt{P$\oplus$*} models performs better than \texttt{W}$_{r+b}$, that is surprising given the success they depict for \To\ (Table~\ref{table:multiclass_task_results}).
However, \texttt{C} with multi-head attention (\texttt{C$\oplus$I$\oplus$J$\oplus$A}) show a significant improvement of 4.6\% over its counterpart, that is very encouraging. % (\texttt{C$\oplus$I$\oplus$J}).

\begin{table}[htbp!]
\centering\resizebox{\columnwidth}{!}{
\begin{tabular}{l||c|c|r}
 & \multicolumn{1}{c|}{\textbf{\texttt{DEV}}} & \multicolumn{1}{c|}{\textbf{\texttt{TST}}} &  \multicolumn{1}{c}{\bm{$\delta$}} \\ 
\hline \hline
\texttt{W$_{r+b}$}                              & 76.24 ($\pm$1.08) & 77.70 ($\pm$0.59) &  \multicolumn{1}{c}{-} \\ 
\hline
\texttt{P$\oplus$I$\oplus$J}                    & 74.73 ($\pm$0.54) & 75.60 ($\pm$1.07) & -2.1 \\ 
\texttt{P$\oplus$I$\oplus$J$\oplus$A}           & 75.36 ($\pm$0.57) & 77.25 ($\pm$0.87) & -0.5 \\ 
\texttt{P$\oplus$I$\oplus$J$\oplus$A$\ominus$E} & 76.42 ($\pm$0.22) & 77.58 ($\pm$0.95) & -0.1 \\ 
\hline
\texttt{C$\oplus$I$\oplus$J}                    & 73.85 ($\pm$0.87) & 74.65 ($\pm$1.87) & -3.1 \\ 
\texttt{C$\oplus$I$\oplus$J$\oplus$A}           & \textbf{76.99} ($\pm$1.10) & \textbf{79.20} ($\pm$0.26) & \textbf{1.5} \\
\texttt{C$\oplus$I$\oplus$J$\oplus$A$\ominus$E} & 76.20 ($\pm$0.96) & 78.49 ($\pm$0.74) & 0.8 \\
\end{tabular}}
\caption{Accuracy ($\pm$ standard deviation) on the development (\texttt{DEV}) and test (\texttt{TST}) sets for \Tt, achieved by the models in Section~\ref{ssec:models}. $\delta$: delta over \texttt{W$_r$} on \texttt{TST}.}
\label{table:binary_class_task_results}
\end{table}

\noindent Multi-head attention (\texttt{A}) gives good improvement to \texttt{P} as well.
Interestingly, the one excluding the embedding list (\texttt{$\ominus$E}) performs slightly better than the one including it (\texttt{P$\oplus$I$\oplus$J$\oplus$A}), implying that the embeddings from the pruned sections are not as useful once the attention is in place.

%%%%%%%%%%%%%%%%%%%%%%%%%%%%%% Analysis %%%%%%%%%%%%%%%%%%%%%%%%%%%%%%

\subsection{Analysis}
\label{ssec:analysis}

\begin{figure}[htbp!]
\centering
\includegraphics[scale=0.235]{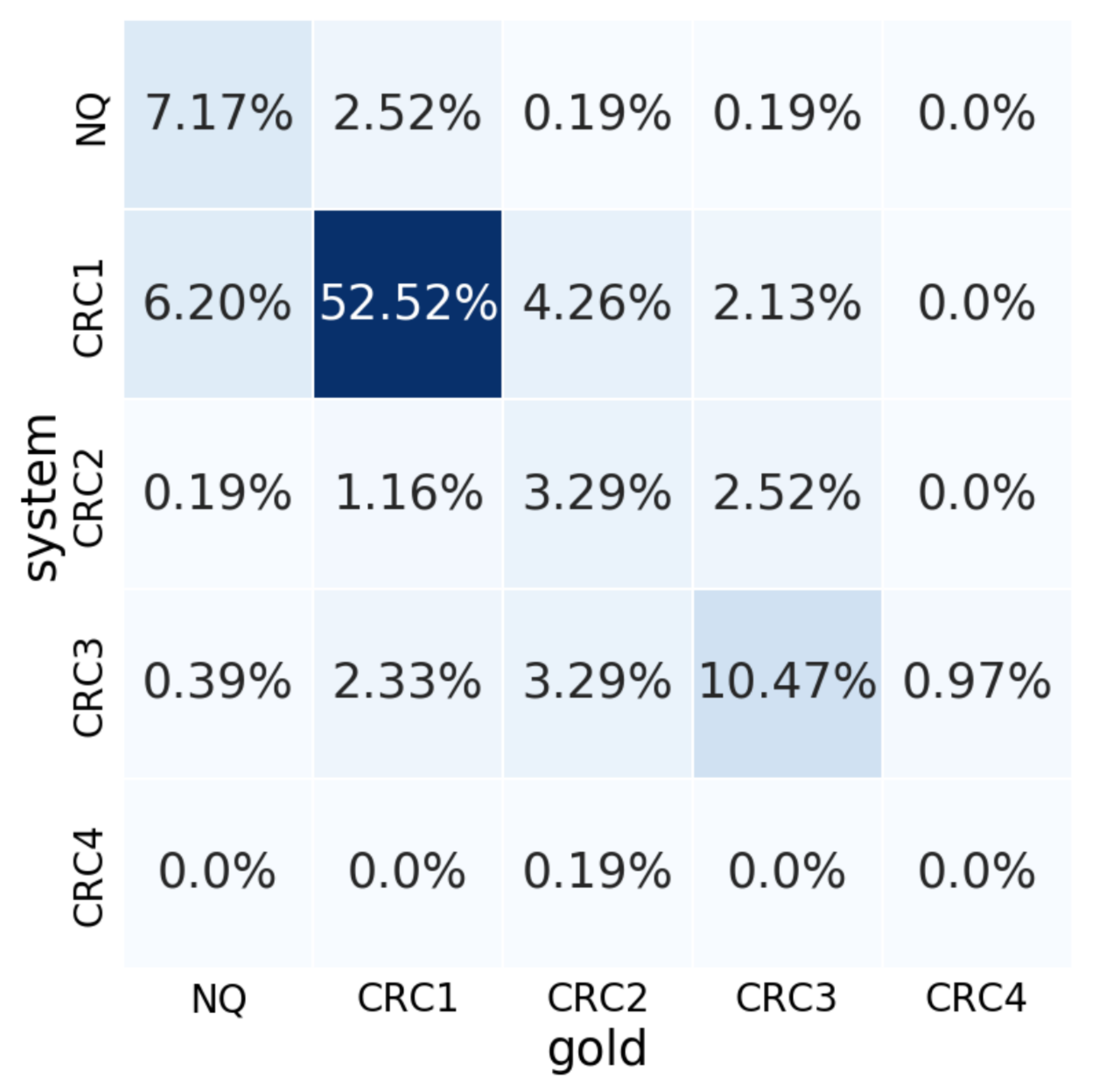}
\caption{Confusion matrix for the best model of \To.}
\label{fig:confusion-matrix-t1}
\end{figure}

Figure~\ref{fig:confusion-matrix-t1} shows the confusion matrix for \To's best model, \texttt{C$\oplus$I}. 
The prediction of \CRCo\ shows robust performance, which has the most number of training instances (Table~\ref{tab:t1-data-split}), whereas the other dimensions are mostly confused around their neighbors, often hard to distinguish even for human experts.

\begin{figure}[htbp!]
\centering
\includegraphics[scale=0.145]{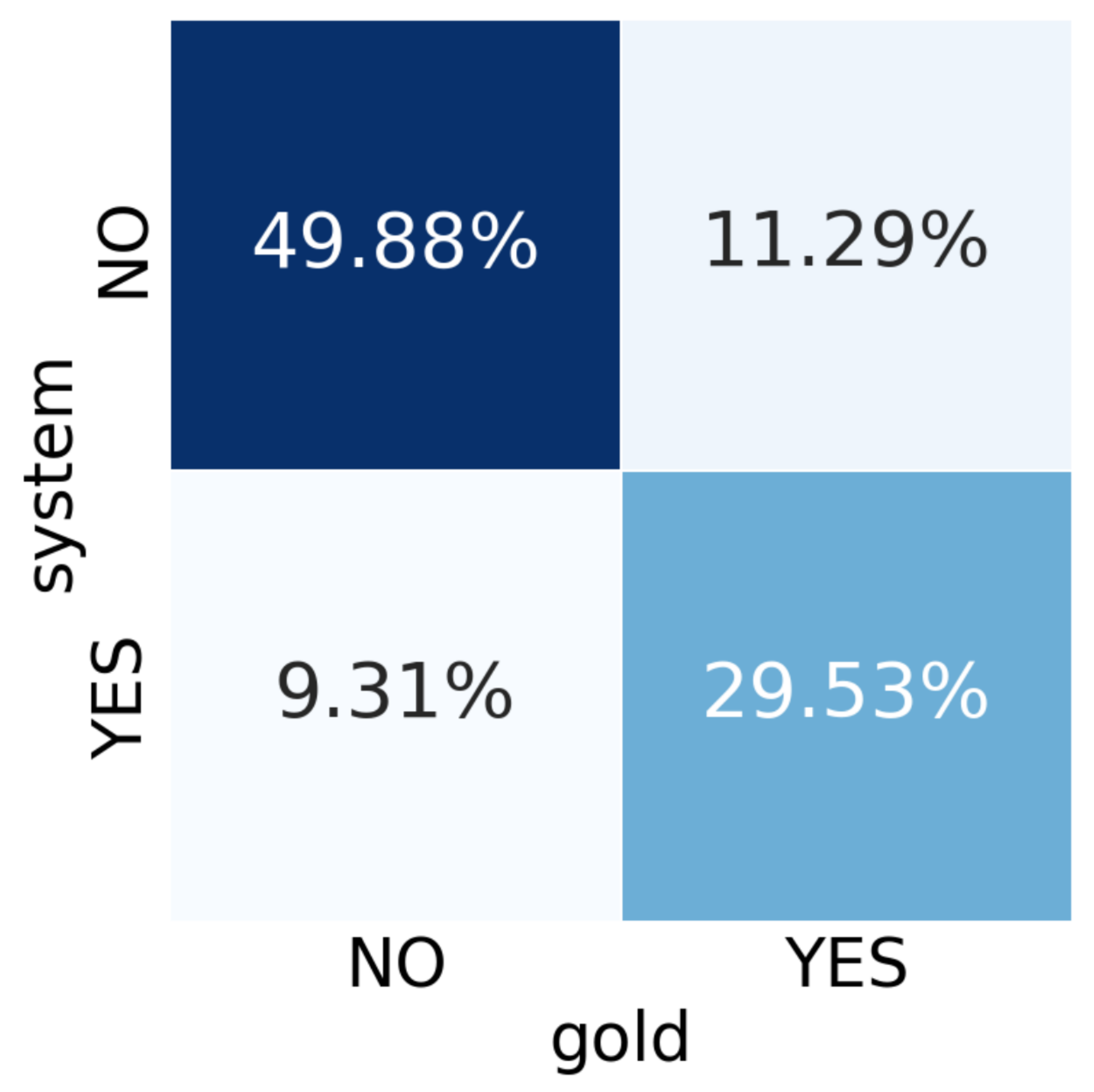}
\caption{Confusion matrix for the best model of \Tt.}
\label{fig:confusion-matrix-t2}
\end{figure}

\noindent Figure~\ref{fig:confusion-matrix-t2} shows the confusion matrix for \Tt's best model, \texttt{C$\oplus$I$\oplus$J$\oplus$A}.
In general, this model shows robust performance across all dimensions.

\subsection{Error Analysis}
\label{ssec:error-analysis}

This section provides a detailed analysis from our experts about prediction errors made by our best model in Section~\ref{ssec:results}.

\paragraph{General} The following observations are found as general error cases:

\begin{itemize}[leftmargin=*]
\setlength\itemsep{0em}
\item Classifying foreign trained MDs and persons with PhDs with no clinical research experience to overrate them. 
(1) It picks up research project done in training as significant research.
(2) It is unable to identify clinical research experience.

\item Classifying laboratory personnel entering CRC area.

\item Counting research experience: identifying dates of experience.
(1) It needs to accumulate experience (e.g., \CRCo: 6 months; \CRCt: 2-3 years). 
(2) It needs implications for creating a structured entry form versus resume 
(3) Academic research experiences that are less than 1000 hours not counted; a semester experience not counted.
(4) It needs to count paid research experience.

\item Not picking up research related titles or terms such as
(1) Research coordinator, research assistant, senior assistant;
(2) IRB, informed consent, regulatory, specimen management, SOP, interviews, questionnaires; %, surveys.
(3) Lab researcher: assays, immunohistochemistry.

\item Not recognizing transferable skills such as clinical setting, clinical experience, and laboratory experience.

\item Recognizing correct certifications.
CRC positions require Clinical Research Certification but do not require CITI or CPR Certificates.

\item Not distinguishing levels of preparation and associated clinical experience or research experience.
Distinguishing scientific vs.\ nonscientific degrees for \CRCo\ and \CRCt\ is particularly important.
\end{itemize}

\noindent The following error cases are found between the adjacent pairs of CRC positions:

\paragraph{\NQ\ vs.\ \CRCo}
It needs to distinguish transferable skills, clinical setting, clinical experiences.

\paragraph{\CRCo\ vs.\ \CRCt}
It needs to count for 
(1) Levels of education;
(2) Scientific vs.\ non-scientific degree;
(3) Clinical experience that is a must for \CRCt\ at lower educational levels;

\paragraph{\CRCt\ vs.\ \CRCr}
It needs to count for 
(1) Length of clinical research experience;
(2) Foreign trained MD;
(3) Laboratory personnel length of time

\paragraph{\CRCr\ vs.\ \CRCf}
(1) Foreign MD are often classified too high.
(2) \CRCf\ needs Certification in Clinical Research

% To be added
\section{Conclusion}
\label{sec:conclusion}

This paper proposes two novel tasks, competence-level classification (\To) and resume-description matching (\Tt), and provides a high-quality dataset as well as robust models using several transformer-based approaches.
The accuracies achieved by our best models, 73.3 for \To\ and 79.2 for \Tt, show a good promise for these models to be deployed in real HR systems.
To the best of our knowledge, this is the first time that those two tasks are thoroughly studies, especially with the latest transformer architectures.
We will continuously explore to improve these models by integrating expert's knowledge.

%Transformer for resume and job description should be different.
%Intead of using sum, we should use concat or other.

\bibliography{emnlp2020}

\begin{thebibliography}{10}
\expandafter\ifx\csname natexlab\endcsname\relax\def\natexlab#1{#1}\fi

\bibitem[{Chifu et~al.(2017)Chifu, Chifu, Popa, and Salomie}]{chifu:17a}
Emil~St. Chifu, Viorica~R. Chifu, Iulia Popa, and Ioan Salomie. 2017.
\newblock \href {https://ieeexplore.ieee.org/document/8117003/} {{A System for
  Detecting Professional Skills from Resumes Written in Natural Language}}.
\newblock In \emph{Proceedings of the IEEE International Conference on
  Intelligent Computer Communication and Processing}, ICCP'17, pages 189--196.

\bibitem[{Deng et~al.(2018)Deng, Lei, Li, and Lin}]{deng:18a}
Yu~Deng, Hang Lei, Xiaoyu Li, and Yiou Lin. 2018.
\newblock \href {https://ieeexplore.ieee.org/document/8396176} {{An Improved
  Deep Neural Network Model for Job Matching}}.
\newblock In \emph{{Proceedings of the International Conference on Artificial
  Intelligence and Big Data}}, ICAIBD'18.

\bibitem[{Devlin et~al.(2019)Devlin, Chang, Lee, and Toutanova}]{devlin_2019}
Jacob Devlin, Ming-Wei Chang, Kenton Lee, and Kristina Toutanova. 2019.
\newblock \href {https://doi.org/10.18653/v1/N19-1423} {{BERT: Pre-training of
  Deep Bidirectional Transformers for Language Understanding}}.
\newblock In \emph{Proceedings of the 2019 Conference of the North {A}merican
  Chapter of the Association for Computational Linguistics: Human Language
  Technologies, Volume 1 (Long and Short Papers)}, pages 4171--4186,
  Minneapolis, Minnesota. Association for Computational Linguistics.

\bibitem[{Myers(2019)}]{myers:19a}
Melanie~A. Myers. 2019.
\newblock \href {https://academic.oup.com/jamia/article/26/5/383/5369358}
  {{Healthcare Data Scientist Qualifications, Skills, and Job Focus: A Content
  Analysis of Job Postings}}.
\newblock \emph{Journal of the American Medical Informatics Association},
  26(5):383--391.

\bibitem[{{Nasser} et~al.(2018){Nasser}, {Sreejith}, and
  {Irshad}}]{nasser_2018}
S.~{Nasser}, C.~{Sreejith}, and M.~{Irshad}. 2018.
\newblock \href {https://ieeexplore.ieee.org/document/8529097} {{Convolutional
  Neural Network with Word Embedding Based Approach for Resume
  Classification}}.
\newblock In \emph{2018 International Conference on Emerging Trends and
  Innovations In Engineering And Technological Research (ICETIETR)}, pages
  1--6.

\bibitem[{Sayfullina et~al.(2017)Sayfullina, Malmi, Liao, and
  Jung}]{sayfullina_2017}
Luiza Sayfullina, Eric Malmi, Yiping Liao, and Alexander Jung. 2017.
\newblock \href {https://link.springer.com/chapter/10.1007/978-3-319-73013-4_8}
  {{Domain adaptation for resume classification using convolutional neural
  networks}}.
\newblock In \emph{International Conference on Analysis of Images, Social
  Networks and Texts}, pages 82--93. Springer.

\bibitem[{Stewart(2019)}]{darin:19a}
Darin Stewart. 2019.
\newblock \href
  {https://www.gartner.com/en/documents/3899769/understanding-your-customers-by-using-text-analytics-and}
  {{Understanding Your Customers by Using Text Analytics and Natural Language
  Processing}}.
\newblock \emph{Gartner Research}, G00373854.

\bibitem[{Valdez-Almada et~al.(2018)Valdez-Almada, Rodriguez-Elias, Rose-Gomez,
  Velazquez-Mendoza, and Gonzalez-Lopez}]{valdez-almada:18a}
Rogelio Valdez-Almada, Oscar~M. Rodriguez-Elias, Cesar~E. Rose-Gomez, Maria
  D.~J. Velazquez-Mendoza, and Samuel Gonzalez-Lopez. 2018.
\newblock \href {https://ieeexplore.ieee.org/document/8337940} {{Natural
  Language Processing and Text Mining to Identify Knowledge Profiles for
  Software Engineering Positions Generating Knowledge Profiles from Resumes}}.
\newblock In \emph{{Proceedings of the International Conference in Software
  Engineering Research and Innovation}}, CONISOFT'18.

\bibitem[{Vaswani et~al.(2017)Vaswani, Shazeer, Parmar, Uszkoreit, Jones,
  Gomez, Kaiser, and Polosukhin}]{vaswani_2017}
Ashish Vaswani, Noam Shazeer, Niki Parmar, Jakob Uszkoreit, Llion Jones,
  Aidan~N. Gomez, Lukasz Kaiser, and Illia Polosukhin. 2017.
\newblock \href {http://dl.acm.org/citation.cfm?id=3295222.3295349} {Attention
  is all you need}.
\newblock In \emph{Proceedings of the 31st International Conference on Neural
  Information Processing Systems}, NIPS'17, pages 6000--6010, USA. Curran
  Associates Inc.

\bibitem[{{Zaroor} et~al.(2017){Zaroor}, {Maree}, and {Sabha}}]{zaroor_2017}
A.~{Zaroor}, M.~{Maree}, and M.~{Sabha}. 2017.
\newblock \href {https://ieeexplore.ieee.org/document/8372026} {{JRC: A Job
  Post and Resume Classification System for Online Recruitment}}.
\newblock In \emph{2017 IEEE 29th International Conference on Tools with
  Artificial Intelligence (ICTAI)}, pages 780--787.

\end{thebibliography}
\bibliographystyle{acl_natbib}

\cleardoublepage\appendix
\section{Appendices}
\label{sec:supplemental-materials}
\subsection{Spliting Algorithm for \Tt}
\label{ssec:splitting_algortim}
Algorithm~\ref{algortim:splitting} is to split the \texttt{TRN}/\texttt{DEV}/\texttt{TST} sets for \Tt\ (Table~\ref{tab:t2-data-split}) without overlapping applicants across them while keeping the label distributions. 
The key idea is to split the data by targeted label distributions but with a smaller training set ratio than the original one. 
If there are overlapping applicants, then it puts all of the overlaps into the training set so that the training set ratio will be large enough to be close to the targeted training set ratio while the label distributions are still kept in a great extent.

\begin{algorithm}[htbp!]
\small
\SetAlgoLined
\KwResult{The splitted dataset for \Tt}
Initialize a random training set ratio $T_i$ smaller than the targeted training and evaluation ratio $T_t$\;
 \While{True}{
    Split the training and evaluation set by $T_i$ based on the ratio $R$ of positions applied and annotated matching results\;
 \eIf{There are overlap resumes between training and evaluation set}{
   Put all overlap resumes into the splitted training set\;
   Compute the new training ratio $T_n$\;
   \eIf{$T_n$ is not closed to $T_t$}
   {
     Adjust $T_i$ based on the relation between $T_n$ and $T_t$\;
     Continue\;
   }
   {
     Split the evaluation set into the development and test set based on $R$\;
     Return the splitted set\;
   }
   }{
   \eIf{$T_i$ is not closed to $T_t$}{
     Adjust $T_i$ based on the relation between $T_i$ and $T_t$\;
     Continue\;
   }{
     Split the evaluation set into the development and test set based on $R$\;
     Return the splitted set\;
   }
   }
  }
 \caption{Splitting Algorithm for \Tt}
 \label{algortim:splitting}
\end{algorithm}
\vspace{-2ex}

\subsection{Experimental Settings}
\label{ssec:experimental-settings}

Table~\ref{table:hyperparameters} shows the hyper-parameters used for each model (Section~\ref{ssec:models}). 
For chunk segmenting in Section~\ref{sssec:multi-class-chunk-segmenting}, let $k_i$ be the number of chunks in the $i$'th section, then $K = \sum_{i=1}^m k_i$ is the total number of chunks in $R$. 
To utilize the GPU memory wisely,\LN resumes with the same $K$ are put to the same batch and different batches are trained with different batch sizes based on $K$ and GPU memory to maximum the GPU usage. 
Different seeds are used when developing models for three times.

\begin{table}[htbp!]
\centering\small\resizebox{\columnwidth}{!}{
\begin{tabular}{c||c|c|c|c|c|c|c}
\textbf{Model} & \textbf{L} & \textbf{GAS} & \textbf{BS}       & \textbf{LR} & \textbf{E} & \textbf{T} & \textbf{PS} \\ \hline \hline
\texttt{W}$_r$/\texttt{W$_{r+b}$} & 512  & 2   & 5       & 2e-05    & 20  & 1-3h  &  109M    \\ \hline
\texttt{P}/\texttt{P$\oplus$I}    & 512  & 2   & 3       & 2e-05    & 20  & 4-6h  &  109M   \\ 
\texttt{C}/\texttt{C$\oplus$I}    & 128  & 1   & 1,2,4   & 2e-05    & 20  & 4-6h  &  109M   \\ \hline
\texttt{P$\oplus$I$\oplus$*}    & 512  & 2    & 3       & 2e-05    & 20   & 6-8h  & 112M     \\    
\texttt{C$\oplus$I$\oplus$*} \ & 128  & 1    & 1,2,4   & 2e-05    & 20  & 6-8h  &  112M    \\
\end{tabular}}
\caption{Hyperparameters. L: \texttt{TE} input length; GAS: gradient accumulation steps; BS: batch size; LR: learning rate; E: number of training epochs; T: approximate training time(h: hours); PS: approximate models training parameters size.}
\label{table:hyperparameters}
\end{table}

\subsection{Analysis on Section Pruning}

Section pruning is used to discard insignificant tokens in order to meet the limit of input size required by the transformer encoder (Section~\ref{sssec:multi-class-section-pruning}).
Tables~\ref{table:section_length_before_pruning} and \ref{table:section_length_after_pruning} show the section lengths before and after section pruning, respectively. 
These tables show that section pruning can noticeably reduce the maximum and average lengths of the sections.

 %, but also after pruning about 99\% of each section content is smaller than the max length which enables transformer models to encode more important tokens.

\begin{table}[htbp!]
\centering\small\resizebox{\columnwidth}{!}{
\begin{tabular}{l||c|c|c}
\textbf{Section} & \textbf{Average ($\pm$stdev)} & \textbf{Max} & \textbf{Ratio} \\ \hline \hline
\texttt{Profile}          & 100.65 ($\pm$215.75)               & 2139 &  94.93\%      \\
\texttt{Skills}           & 60.70 ($\pm$102.61)                & 1157 &  98.95\%              \\
\texttt{Work Experience}  & 314.61 ($\pm$316.61)               & 3605 &  80.26\%              \\
\texttt{Education}        & 174.30 ($\pm$289.37)               & 3662 &  89.50\%              \\
\texttt{Other}            & 77.41 ($\pm$145.40)                & 2184 &  98.34\%              \\
\texttt{Activities}       & 168.09 ($\pm$289.40)               & 3967 &  91.13\%             
\end{tabular}}
\caption{Section lengths before section pruning (Section~\ref{sssec:multi-class-section-pruning}). Average/Max: the average and max lengths of input sections. Ratio: the ratios of input sections that are under the max-input length restricted by the transformer encoder.}
\label{table:section_length_before_pruning}
\end{table}

\begin{table}[htbp!]
\centering\small\resizebox{\columnwidth}{!}{
\begin{tabular}{l||c|c|c}
\textbf{Section} & \textbf{Average($\pm$stdev)} & \textbf{Max} &\textbf{Ratio} \\ \hline \hline
\texttt{Profile} & 77.95($\pm$127.70)       & 1514 &   99.60\%             \\
\texttt{Skills}          & 55.59 ($\pm$70.36)                 & 546  &   99.93\%             \\
\texttt{Work Experience} & 232.63 ($\pm$168.84)               & 2099 &   98.98\%             \\
\texttt{Education}       & 129.91 ($\pm$165.06)               & 1755 &   98.81\%             \\
\texttt{Other}           & 72.19 ($\pm$108.80)                & 1468 &   99.38\%             \\
\texttt{Activities}      & 125.71 ($\pm$57.74)                & 1514 &   99.13\%            
\end{tabular}}
\caption{Section lengths after section pruning (Section~\ref{sssec:multi-class-section-pruning}). Average/Max: the average and max lengths of input sections. Ratio: the ratios of input sections that are under the max-input length restricted by the transformer encoder.}
\label{table:section_length_after_pruning}
\end{table}

\end{document}